\documentclass[letterpaper, 10 pt, conference]{ieeeconf}  

\IEEEoverridecommandlockouts                              
\IEEEoverridecommandlockouts                              

\usepackage{graphicx} 
\usepackage{epsfig} 
\usepackage{amsmath} 
\usepackage{amssymb}  
 \usepackage[colorlinks=true, allcolors=blue]{hyperref}
\usepackage{cite}

\usepackage{xcolor,cleveref}
\usepackage{makecell}
\usepackage{svg}
\crefname{equation}{Eq.}{Eqs.}
\Crefname{equation}{Equation}{Equations}
\usepackage[ruled]{algorithm2e}

\usepackage{amsthm}
\SetKwInput{KwData}{Input}
\SetKwInput{KwResult}{Output}
\LinesNumbered

\newtheorem{theorem}{Theorem}[section]
\theoremstyle{definition}

\newtheorem{problem}[theorem]{Problem}
\newtheorem{lemma}[theorem]{Lemma}
\title{\LARGE \bf
REVISE: Robust Probabilistic Motion Planning\\%
in a Gaussian Random Field
}

\author{Alex Rose$^{1, 3}$, Naman Aggarwal$^{1}$, Christopher Jewison$^{2}$, and Jonathan P. How$^{1}$
\thanks{$^{1}$Aerospace Controls Laboratory, Massachusetts Institute of Technology,
Cambridge, MA, USA. e-mail:
        {\tt\small \{ameredit, namanagg, jhow\}@mit.edu}. This work was supported by the National Science Foundation Graduate Research Fellowship under grant no. 2141064.}%
\thanks{$^{2}$Draper, Cambridge, MA. e-mail: {\tt\small cjewison@draper.com}.}%
\thanks{$^{3}$Draper Scholar, Draper, Cambridge, MA. The authors would like to thank the Draper Scholars program for supporting this work.}
}

\begin{document}
\setlength{\textfloatsep}{0pt}

\maketitle
\thispagestyle{empty}
\pagestyle{empty}

\begin{abstract}
This paper presents Robust samplE-based coVarIance StEering (REVISE), a multi-query algorithm that generates robust belief roadmaps for dynamic systems navigating through spatially dependent disturbances modeled as a Gaussian random field. Our proposed method develops a novel robust sample-based covariance steering edge controller to safely steer a robot between state distributions, satisfying state constraints along the trajectory. Our proposed approach also incorporates an edge rewiring step into the belief roadmap construction process, which provably improves the coverage of the belief roadmap. 
When compared to state-of-the-art methods \cite{ridderhof2022chance, aggarwal2024sdp}, REVISE improves median plan accuracy (as measured by Wasserstein distance between the actual and planned final state distribution) by 10x in multi-query planning and reduces median plan cost (as measured by the largest eigenvalue of the planned state covariance at the goal) by 2.5x in single-query planning for a 6DoF system. We will release our code at \url{https://acl.mit.edu/REVISE/}. 
\end{abstract}

\section{INTRODUCTION}
Robots navigating in complex and uncertain environments often rely on ``roadmaps'' of dynamically feasible trajectories through the environment. Roadmaps are typically graph-structured, with nodes corresponding to system states and edges corresponding to feasible trajectories between states \cite{kavraki1996probabilistic, kuffner2000rrt}. Pre-computing roadmaps offline allows robots to quickly plan paths to new goal states by extending the roadmap to reach a new goal node, then searching for a trajectory through the roadmap to the new goal. Two important qualities in a roadmap are \textit{accuracy}, or the probability that trajectories in the roadmap are dynamically feasible under real environmental conditions, and \textit{coverage}, or the portion of the state space that is reachable from the roadmap.

Many motion planning algorithms incrementally build a rapidly exploring random tree (RRT) of reachable states forward from an initial state \cite{kuffner2000rrt, karaman2011incremental, luders2010chance, luders2013robust}. The RRT* algorithm \cite{karaman2011incremental} 
includes an ``edge rewiring'' step when nodes are added, resulting in an asymptotically optimal tree that preserves the lowest-cost paths from root to leaf throughout construction. CC-RRT \cite{luders2010chance} and CC-RRT* \cite{luders2013robust} extend RRT and RRT* to chance-constrained stochastic systems. CC-RRT and CC-RRT* begin with an initial state distribution and grow a stochastic tree by simulating open-loop control trajectories, enforcing chance constraints along each edge \cite{luders2010chance, luders2013robust}.

 \begin{figure}[thpb]
      \centering
      \includegraphics[width=\columnwidth]{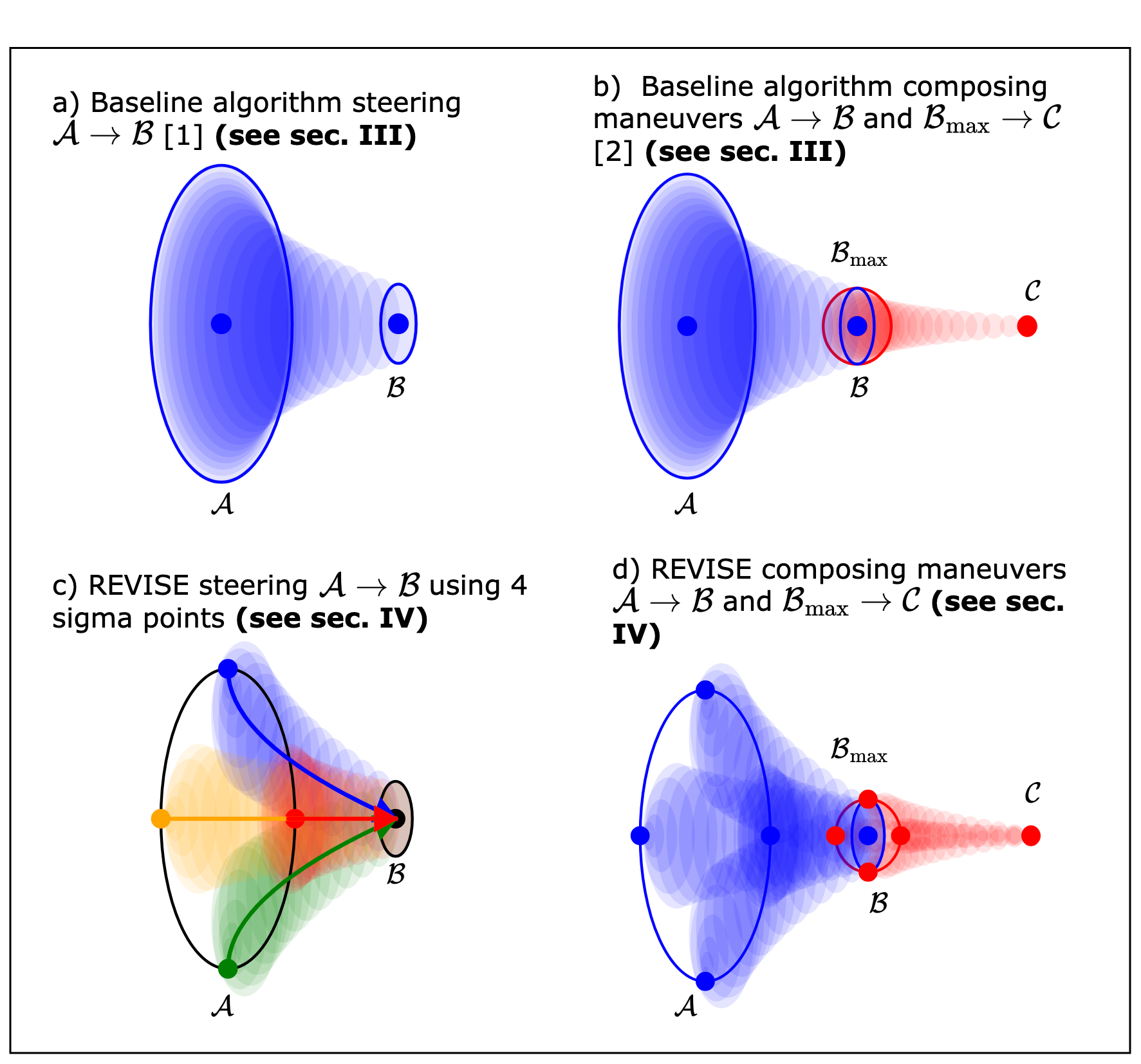}
      \setlength{\abovecaptionskip}{-15pt}
      \caption{(a-b) Baseline covariance steering steers between Gaussian distributions \cite{ridderhof2022chance, aggarwal2024sdp, zheng2024cs}. (c-d) REVISE samples points on a Gaussian distribution, then steers a mixture of Gaussian distributions to a Gaussian distribution. }
      \vspace{0.1cm}
      \label{fig: sigma_vis}
\end{figure}

Other works on belief roadmapping for deterministic systems construct invariant ``funnels'' from feedback controllers, and build a roadmap with funnels as edges. Tedrake et al. \cite{tedrake2010lqr} grow a tree backwards from a goal state and compute funnels corresponding to regions of attraction for each edge in the tree. Then, if a robot starts anywhere inside the ``mouth'' of a funnel, its trajectory is guaranteed to remain within the funnel under the given feedback control policy. Later work \cite{majumdar2017funnel} extends this approach to multi-query planning by computing a library of funnels offline, then sequentially composing funnels online such that the tail of each funnel sits inside the mouth of the next funnel.

Prior works \cite{agha2014firm, zheng2024cs, aggarwal2024sdp} on multi-query belief roadmapping for stochastic systems typically focus on linear (or nearly linear) systems with additive Gaussian noise. FIRM \cite{agha2014firm} builds a roadmap in the belief space, constructing edges using stationary linear quadratic Gaussian (SLQG) control. FIRM relies on a stationary controller and only considers stationary belief nodes. CS-BRM \cite{zheng2024cs} allows non-stationary belief nodes and uses finite-time covariance steering control to construct edges in the roadmap, but does not allow for chance constraints. Related work \cite{ridderhof2022chance} does not address roadmap construction, but provides a framework for chance-constrained covariance steering under spatially dependent disturbances. Our prior work \cite{aggarwal2024sdp} addresses chance constraints and introduces the notion of a maximal-coverage roadmap, where edges are constructed in a way that maximizes the coverage of the roadmap. 

We focus on planning with feedback control for systems with nontrivial nonlinearities resulting from state-dependent disturbances. We develop an algorithm for constructing a belief roadmap for a system disturbed by a state-dependent Gaussian random field. Our contributions are:
\begin{enumerate}
\item A novel robust algorithm for covariance steering in a state-dependent Gaussian random field. This algorithm approximates the state distribution by a set of sigma points subject to different disturbances and steers the state covariance with a novel robust objective that minimizes worst-case state error over all sigma points. 
\item An edge rewiring procedure that provably improves roadmap coverage without adding extra nodes or edges to the roadmap. 
\item Experiments on a 6DoF model where REVISE outperforms state-of-the-art methods, improving median plan accuracy by 10x in multi-query planning and reducing median plan cost by 2.5x in single-query planning. 
\end{enumerate}
\section{PROBLEM STATEMENT}
Consider a discrete-time system with dynamics
\begin{equation}\label{eq:dynamics}
\mathbf{x}_{k+1} = A\mathbf{x}_k + B\mathbf{u}_k + G\Psi(\phi(\mathbf{x}_k)),
\end{equation}
such that the system has linear dynamics that are disturbed by a Gaussian random field $\Psi(\cdot)$, which has a dependent variable $\phi$ that is a function of the state $\mathbf{x}_k$.

We define a generic finite-horizon optimal control problem where an N-step steering maneuver is taken from initial distribution $\mathcal{I}$ to goal distribution $\mathcal{G}$ in Problem \ref{prob: opt_steer_grf}, subject to the dynamics in Equation \ref{eq:dynamics}.
\begin{problem} \label{prob: opt_steer_grf}
\textit{Steer from initial distribution $\mathcal{I}$ to goal distribution $\mathcal{G}$, over $N$ steps, subject to controller parameterization $f_k(\cdot)$ such that $\mathbf{u}_k = f_k(\mathbf{x}_k)$ and cost $c_k(\mathbf{u}_k, \mathbf{x}_{k+1})$.}
\begin{equation}
\min_{f_k} J = \mathbb{E}\left[ \sum_{k=0}^{N-1} c_k(\mathbf{u}_k, \mathbf{x}_{k+1}) \right]
\end{equation}
such that:
\begin{equation}
\begin{split}
&\mathbf{x}_{k+1} = A\mathbf{x}_k + B\mathbf{u}_k + G\Psi(\phi(\mathbf{x}_k)) \\
&\mathbf{x}_0 \sim \mathcal{N}(\mu_\mathcal{I}, \Sigma_\mathcal{I}),\ \mathbf{x}_N \sim \mathcal{N}(\mu_N, \Sigma_N) \\
&\mu_N = \mu_\mathcal{G},\ \Sigma_N \preceq \Sigma_\mathcal{G} \\
&\mathbb{P}(\mathbf{x}_k \in \mathcal{X}) \geq 1 - \epsilon_x, \ \mathbb{P}(\mathbf{u}_k \in \mathcal{U}) \geq 1 - \epsilon_u.
\end{split}
\end{equation}
\end{problem}
However, it may be impossible to directly steer from $\mathcal{I}$ to $\mathcal{G}$ in $N$ steps (e.g. due to control constraints or obstacles). We address the challenge of finding a feasible path to $\mathcal{G}$, even if $> N$ steps are required, by building a belief roadmap that can \textit{compose} multiple $N$-step maneuvers beginning at $\mathcal{I}$ and eventually reaching $\mathcal{G}$. Each node in the roadmap corresponds to a Gaussian distribution in the state space, and each edge in the roadmap corresponds to a control policy that steers between nodes, found by an \textit{edge controller} of the same form as Problem \ref{prob: opt_steer_grf}. The roadmap can be reused for multi-query planning, steering from $\mathcal{I}$ to different goal distributions reachable from nodes in the roadmap.

Our prior work \cite{aggarwal2024sdp} presents an algorithm for building a belief roadmap \textit{backward} from a goal distribution $\mathcal{G}$ under white noise. In this paper we present an algorithm for building a belief roadmap \textit{forward} from an initial distribution $\mathcal{I}$ under a Gaussian random field, solving the following:
\begin{problem}\label{prob: feasibility}
\textit{Find paths from the initial distribution $\mathcal{I}$ to all goal distributions $\mathcal{G}$ for which paths exist (subject to state and control constraints).}
\end{problem}

\section{PRELIMINARIES: COVARIANCE STEERING IN A GAUSSIAN RANDOM FIELD}\label{sec: cov_steer_background}
We use the notation that $\mathbf{V}$ represents a column of stacked $\mathbf{v}_k$ $\forall k=0,\ldots, N$, $\overline{\mathbf{V}}$ is the mean of $\mathbf{V}$, and $\widetilde{\mathbf{V}} = \mathbf{V}-\overline{\mathbf{V}}$. We also use $\mathbf{w}_k = \Psi(\phi(\mathbf{x}_k))$, such that $\mathbf{W}$ represents a column of stacked $\Psi(\phi(\mathbf{x}_k))$ $\forall k=0, \ldots, N$.
%
The state dynamics from \cref{eq:dynamics} propagated over  $\forall k=0, \ldots, N$ can be written in block-matrix notation as \cite{ridderhof2022chance, okamoto2018optimal, bakolas2016optimal, skaf2010design}:
\begin{equation}
\mathbf{X} = \check{A}\mathbf{x}_0 + \check{B}\mathbf{U} + \check{G}\mathbf{W}
\end{equation}
with the rows of $\check{A}, \check{B},$ and $\check{G}$ given by the fact that $\forall k$,
\begin{equation}
\mathbf{x_k} = A^{k}\mathbf{x}_0 + \sum_{i=0}^k A^{(k-1-i)}B\mathbf{u}_i + \sum_{i=0}^k A^{(k-1-i)}G\mathbf{w}_i.
\end{equation}
$\mathbf{U}$ is given by the state history feedback law
\begin{equation}\label{eq:state_history_feedback}
    \mathbf{u_k} = \sum_{i=0}^k K_{k, i}(\mathbf{x_i} - \overline{\mathbf{x}}_i) + \mathbf{v_k},
\end{equation}
with $\overline{\mathbf{x}}_i$ as the mean value of $\mathbf{x_i}$ \cite{ridderhof2022chance}. The Markov assumption does not hold when disturbances are given by a Gaussian random field (e.g. if $\mathbf{x}_k = \mathbf{x}_{k+n}$ for some $n > 0$, revisiting the state $\mathbf{x}_k$ at time $k + n$ may lead to information gain about the disturbance experienced at time $k$). State history feedback allows the controller to account for the impact of the latent disturbance history on the current state. In block-matrix notation, the control law is
$\mathbf{U} = K\widetilde{\mathbf{X}} + \mathbf{V},$
and the closed-loop state dynamics, mean dynamics, and error-state dynamics are 
\begin{align}
\mathbf{X} &= \check{A}\mathbf{x_0} + \check{B}K\widetilde{\mathbf{X}} + \check{B}\mathbf{V} + \check{G}\mathbf{W}, \notag \\
\label{eq:mean_state_dynamics}
    \overline{\mathbf{X}} &= \check{A}\overline{\mathbf{x_0}} + \check{B}\mathbf{V} + \check{G}\overline{\mathbf{W}}, \\
\widetilde{\mathbf{X}} &= \check{A}\widetilde{\mathbf{x_0}} + \check{B}K\widetilde{\mathbf{X}} + \check{G}\widetilde{\mathbf{W}} = (I - \check{B}K)^{-1}(\check{A}\widetilde{\mathbf{x}}_0 + \check{G}\widetilde{\mathbf{W}}). \notag
\end{align}
Suppose the Gaussian random field $\Psi(\cdot)$ has a mean function $\overline{\Psi}(\phi(\mathbf{x}))$ and a covariance function $\Sigma_\Psi(\phi(\mathbf{x}_i), \phi(\mathbf{x}_j))$. Following \cite{ridderhof2022chance}, given an initial guess control sequence $\hat{\mathbf{U}}$ and an initial state mean $\mu_\mathcal{I}$, we propagate the nominal state trajectory $\hat{\mathbf{X}}$ and disturbance statistics to roll out
$\hat{\mathbf{X}} = \check{A}\mu_\mathcal{I} + \check{B}\hat{\mathbf{U}} + \check{G}\overline{\Psi}(\phi(\hat{\mathbf{X}})).$ Then, we discretize the Gaussian random field $\Psi(\cdot)$ around the nominal trajectory $\hat{\mathbf{X}}$ such that $\overline{\mathbf{W}} = \overline{\Psi}(\phi(\mathbf{\hat{X}}))$ and $\Sigma_\mathbf{W} = \mathbb{E}[\widetilde{\mathbf{W}}\widetilde{\mathbf{W}}^T]$. With this approximation, $\overline{\mathbf{W}}$ and $\Sigma_\mathbf{W}$ are both constant, and the disturbance $\mathbf{W}$ from the Gaussian random field can be approximately modeled by $\mathcal{N}(\overline{\mathbf{W}}, \Sigma_\mathbf{W})$, with no dependence on the state trajectory $\mathbf{X}$ \cite{ridderhof2022chance}.

As in \cite{skaf2010design, ridderhof2022chance}, we define a new decision variable $L = K(I-\check{B}K)^{-1}$. Then, the closed-loop state and control covariance are $\Sigma_\mathbf{X} = \mathbb{E}[\widetilde{\mathbf{X}}\widetilde{\mathbf{X}}^T] = (I + \check{B}L)S(I+\check{B}L)^T$ and
$\Sigma_\mathbf{U} = \mathbb{E}[\widetilde{\mathbf{U}}\widetilde{\mathbf{U}}^T] = LSL^T,$ where $S = \check{A}\Sigma_0\check{A}^T + \check{G}\Sigma_\mathbf{W}\check{G}^T$ is the open-loop state covariance (e.g. if $K = 0$, $\mathbb{E}[\widetilde{\mathbf{X}}\widetilde{\mathbf{X}}^T] = S$), with $\Sigma_0$ equal to the initial state covariance.

We consider polytopic state and control constraints of the form $\mathcal{X} := \{\mathbf{x}_k \in \mathbb{R}^{n} | \cap_{j=1}^{M_x} \alpha_{x, j}^T\mathbf{x}_k \leq \beta_{x, j}\}$ and $\mathcal{U} := \{\mathbf{u}_k \in \mathbb{R}^m | \cap_{j=1}^{M_u} \alpha_{u, j}^T\mathbf{u}_k \leq \beta_{u, j}\}$, which can be converted to chance constraints
$\mathbb{P}(\alpha_{x, j}^T\mathbf{x}_k \leq \beta_{x, j}) \geq 1 - \epsilon_{x, j}$ and
$\mathbb{P}(\alpha_{u, i}^T\mathbf{u}_k \leq \beta_{u, i}) \geq 1 - \epsilon_{u, i}$
for every $k = 0,\ldots,N$, $j = 1,\ldots M_x$, and $i = 1, \ldots M_u$. Following \cite{ridderhof2022chance, okamoto2018optimal}, $\mathbf{x}_k$ is a Gaussian random variable with mean $E_k^x\overline{\mathbf{X}}$ and covariance $E_k^x \Sigma_\mathbf{X} {E_k^x}^T$, where $E_k^x \in \mathbb{R}^{m\times Nm}$ and $E_k^x \mathbf{X} = \mathbf{x}_k$. Then, 
$$\mathbb{P}(\alpha_{x, j}^T \mathbf{x}_k \leq \beta_{x, j}) = \Phi\left(\frac{\beta_{x, j}-\alpha_{x, j}^TE_k^x\overline{\mathbf{X}}}{\sqrt{\alpha_{x, j}^TE_k^x\Sigma_\mathbf{X}E_k^{x^T}\alpha_{x, j}}} \right),$$
where $\Phi(\cdot)$ is the normal cumulative distribution function. With $S^{1/2}$ such that $S^{1/2}(S^{{1/2}})^T = S$ and $S^{T^{1/2}} = (S^{{1/2}})^T$, the state chance constraints can be reformulated exactly in convex form for every $k=0, \ldots, N$, $j = 1, \ldots, M_x$ by \cite{okamoto2018optimal}
$$\beta_{x, j} \geq \alpha_{x, j}^TE_k^x\overline{\mathbf{X}} + \Phi^{-1}(1-\epsilon_{x, j})||S^{T^{1/2}}(I + \check{B}L)^TE_k^{x^T}\alpha_{x, j}||,$$
and the control chance constraints for every $k=0, \ldots, N-1$, $i = 1, \ldots, M_u$ are equivalently reformulated as
$$\beta_{u, i} \geq \alpha_{u, i}^TE_k^u\mathbf{V} + \Phi^{-1}(1-\epsilon_{u, i})||S^{T^{1/2}}L^TE_k^{x^T}\alpha_{u, i}||.
$$

 In our prior work \cite{aggarwal2024sdp}, we demonstrated that when steering \textit{backward} from a goal distribution $\mathcal{G}$, roadmap coverage is maximized by maximizing the spectral radius of the initial distribution $\mathcal{I}$ and composing funnels such that for two funnels $\mathcal{I}_1 \to \mathcal{G}_1$ and $\mathcal{I}_2 \to \mathcal{G}_2$, in order to have the tail of the first funnel end inside the mouth of the second, $\mu_{\mathcal{G}_1} = \mu_{\mathcal{I}_2}$ and $\lambda_{\min}(\mathcal{I}_2) \geq \lambda_{\max}(\mathcal{G}_1)$. Accordingly, for \textit{forward} steering, coverage is maximized by minimizing the spectral radius of the goal distribution $\mathcal{G}$ and composing funnels as in our prior work \cite{aggarwal2024sdp}. The task of steering from initial distribution $\mathcal{I}$ to goal mean $\mu_\mathcal{G}$ in a Gaussian random field while maximizing roadmap coverage is formalized as a convex problem in Problem \ref{prob: min_cov_grf}.
\begin{problem}\label{prob: min_cov_grf}
\textit{Steer from initial distribution $\mathcal{I}$ to a goal distribution centered at mean $\mu_\mathcal{G}$ over $N$ steps, minimizing the spectral radius of the goal distribution in order to maximize roadmap coverage.}
\begin{equation}
    \min_{\mathbf{V}, L} J = ||S^{T^{1/2}}(I + \check{B}L)^T E_N^T||_2
\end{equation}
\textit{subject to:}
\begin{align} \label{cov_steer_grf_constraints}
\overline{\mathbf{X}} &= \check{A}\mu_\mathcal{I} + \check{B}\mathbf{V} + \check{G}\overline{\mathbf{W}} \notag \\
S &= \check{A}\Sigma_{\mathcal{I}}\check{A}^T + \check{G}\Sigma_\mathbf{W}\check{G}^T, ~~~~ 
E_N\overline{\mathbf{X}} = \mu_\mathcal{G}, \\
\beta_{x, j} &\geq \alpha_{x, j}^TE_k^x\overline{\mathbf{X}} + \Phi^{-1}(1-\epsilon_{x, j})||S^{T^{1/2}}(I + \check{B}L)^TE_k^{x^T}\alpha_{x, j}|| \notag\\
\beta_{u, i} &\geq \alpha_{u, i}^TE_k^u\mathbf{V} + \Phi^{-1}(1-\epsilon_{u, i})||S^{T^{1/2}}L^TE_k^{x^T}\alpha_{u, i}|| \notag
\end{align}
\textit {for all} $k = 0, \ldots, N$, $i = 1, \ldots, M_u$, $j = 1, \ldots, M_x$,
\textit{with}
$
\Sigma_\mathcal{G} = I ||S^{T^{1/2}}(I+\check{B}L)^T E_N^T||^2_2.$
\end{problem}
In Problem \ref{prob: min_cov_grf}, $\Sigma_\mathcal{G}$ is defined such that $\Sigma_\mathcal{G} = \Sigma_{\mathcal{G}_1}$ satisfies the funnel composition constraint $||S^{T^{1/2}}(I+\check{B}L)^T E_N^T||^2_2 \leq {\lambda_{\min}(\Sigma_{\mathcal{G}_1})}$, and that for any $\Sigma_{\mathcal{G}_1}$ satisfying the funnel composition constraint, $\Sigma_{\mathcal{G}_1} \succeq \Sigma_\mathcal{G}$.
\section{ROBUST SIGMA POINT METHOD FOR COVARIANCE STEERING}\label{sec: sigma_pt_method}
For systems with nonlinearities, such as disturbances given by a state-dependent Gaussian random field, errors in linearization can accumulate over a trajectory. Suppose that $\overline{\Psi}(\phi(\mathbf{x}))$ varies significantly for different values of $\phi(\mathbf{x})$. Then, for an initial state $\mathbf{x}_0$ drawn from an initial state distribution $\mathcal{N}(\mu_\mathcal{I}, \Sigma_\mathcal{I})$, $\overline{\Psi}(\phi(\mu_\mathcal{I}))$ may differ greatly from $\overline{\Psi}(\phi(\mathbf{x}_0))$.
Our robust method for covariance steering accounts for this variation by approximating the initial state distribution by a collection of sigma points (see Figure \ref{fig: sigma_vis}).

We select $4n$ sigma points, corresponding to $2n$ initial states symmetrically distributed on the $\sqrt{n}$th covariance contour, where $n$ is the dimensionality of the state space. We use each initial state $\mathbf{x}_0^{(i)}$ with two different approximations of the Gaussian random field $\Psi(\phi(\mathbf{x}))$: one linearized around the mean nominal trajectory $\hat{\mathbf{X}}$, and one linearized around $\hat{\mathbf{X}}^{(i)}$, where $\hat{\mathbf{X}}^{(i)} = \check{A}\mathbf{x}_0^{(i)} + \check{B}\hat{\mathbf{U}} + \check{G}\overline{\Psi}(\phi(\hat{\mathbf{X}}^{(i)}))$.  When feedback control is applied, each sigma point will be steered towards the mean trajectory, so linearizing around both $\hat{\mathbf{X}}$ and $\hat{\mathbf{X}}^{(i)}$ provides a good first-order approximation of the expected disturbance from the Gaussian random field.

REVISE relies on a robust objective that minimizes the largest eigenvalue of the worst-case contribution to the second moment of the final state distribution over all sigma points. Intuitively, our proposed approach seeks to maximize \textit{coverage} of the roadmap while retaining robustness across sigma points in order to improve \textit{accuracy}. The worst-case contribution to the second moment of the final state distribution over all sigma points also serves as an upper bound on the largest eigenvalue of the second moment of the final state distribution. With $4n$ sigma points $(\mathbf{x}_0^{(1)}, \mathbf{W}^{(1)})$, $\ldots (\mathbf{x}_0^{(4n)}, \mathbf{W}^{(4n)})$, the robust objective is
\begin{equation}\label{eq: robust_obj}
\min_{\mathbf{V}, L} \max_i \lambda_{\max}(E_N^x(I + \check{B}L)S^{(i)}(I + \check{B}L)^TE_N^{x^T}),
\end{equation}
where
\begin{align}
S^{(i)} &= \check{G}\Sigma_{\mathbf{W}^{(i)}}\check{G}^T + (\check{A}(\mathbf{x}_0^{(i)} - \overline{\mathbf{x}}_0) + \check{G}(\overline{\mathbf{W}}^{(i)} - \overline{\mathbf{W}}))
\notag \\
&~~~~(\check{A}(\mathbf{x}_0^{(i)} - \overline{\mathbf{x}}_0) + \check{G}(\overline{\mathbf{W}}^{(i)} - \overline{\mathbf{W}}))^T.
\end{align}
We formalize a robust semidefinite problem with the robust objective from Equation \ref{eq: robust_obj} in Problem \ref{prob: robust_cov_grf}.
\begin{problem} \label{prob: robust_cov_grf}
\textit{Steer from initial distribution $\mathcal{I}$ to a goal distribution centered at mean $\mu_\mathcal{G}$ over $N$ steps, minimizing the largest eigenvalue of the second moment of the final state distribution over all sigma points.}
\begin{equation}
    \min_{\mathbf{V}, L} J = \lambda_{\max}(\Sigma_f)
\end{equation}
\textit{subject to (\ref{cov_steer_grf_constraints}) and}
\begin{align} \label{eq: semidefinite_constraints}
E_N^x\overline{\mathbf{X}} &= \sum_{i=1}^{4n} \frac{1}{4n} E_N^x\overline{\mathbf{X}}^{(i)}\!\! = \mu_\mathcal{G}\\ \notag
\forall i, 0 &\preceq \begin{bmatrix} I & S^{(i)^{T^{1/2}}}(I + \check{B}L)^T E_N^{x^T} \\ E_N^x(I + \check{B}L)S^{(i)^{1/2}} & \Sigma_f \end{bmatrix}
\end{align}
\textit{with}
$
\Sigma_\mathcal{G} = I \lambda_{\max}(\Sigma_f).
$
\end{problem}
The semidefinite constraints given in \cref{eq: semidefinite_constraints} ensure that $\Sigma_f \succeq M^{(i)}$ for all $i$, with $M^{(i)} = E_N^x(I + \check{B}L)S^{(i)}(I + \check{B}L)^TE_N^{x^T}$, so $\lambda_{\max}(\Sigma_f) \geq \lambda_{\max}(M^{(i)})$ for all $i$. Because the objective minimizes $\lambda_{\max}(\Sigma_f)$, at optimum we will have that $\lambda_{\max}(\Sigma_f) = \max_i \lambda_{\max}(M^{(i)})$, and so the semidefinite relaxation is lossless. $\Sigma_f$ conservatively over-approximates the second moment of the final state distribution. $\Sigma_\mathcal{G}$ is defined such that $\Sigma_\mathcal{G} = \Sigma_{\mathcal{G}_1}$ satisfies the funnel composition constraint $||\Sigma_f||_2^2 \leq \lambda_{\min}(\Sigma_{\mathcal{G}_1})$ and for any $\Sigma_{\mathcal{G}_1}$ satisfying the above constraint, $\Sigma_{\mathcal{G}_1} \succeq \Sigma_\mathcal{G}$.
\section{BELIEF ROADMAP CONSTRUCTION}\label{sec: brm}
We solve Problem \ref{prob: feasibility} by constructing a belief roadmap starting at an initial distribution $\mathcal{I}$, where all nodes in the roadmap represent state distributions reachable from $\mathcal{I}$. This tree-structured roadmap $\mathcal{T}$ is represented by a set of nodes $\mathcal{V}$ and a set of edges $\mathcal{E}$. Each node $v_i \in \mathcal{V}$ represents a state distribution with mean $\mu_i$ and covariance $\Sigma_i$. Each edge $e_{i \to j} \in \mathcal{E}$ represents an edge controller between nodes $v_i$ and $v_j$, and includes pointers to $v_i$ and $v_j$, feedback control gain $\mathbf{K}_{i \to j}$, and open-loop control $\mathbf{V}_{i \to j}$. 

Our prior work \cite{aggarwal2024sdp} uses a belief roadmap construction procedure similar to that given in Algorithm \ref{alg: no_rewiring}. First, a node $v_k$ is randomly selected from the node set $\mathcal{V}$ according to the Voronoi bias of the node means. Then, a query mean $\mu_q$ is sampled from a region around the node $v_k$, and a steering maneuver is attempted from $v_k$ to $\mu_q$ with an edge controller $\Pi$ (e.g. Problem \ref{prob: min_cov_grf} or \ref{prob: robust_cov_grf}). If the maneuver is successful, a new node and edge are added to the graph. This procedure randomly samples node means in a similar manner to RRT \cite{kuffner2000rrt}, and then expands nodes in a way that maximizes reachability from the initial state distribution $\mathcal{I}$.
\vspace{-0.4cm}
\begin{algorithm}[ht]\label{alg: no_rewiring}
\caption{Belief roadmap construction without edge rewiring (adapted from our prior work \cite{aggarwal2024sdp})}
$\mathcal{V} \gets \{(\mu_\mathcal{I}, \Sigma_\mathcal{I})\}, \mathcal{E} \gets \emptyset$

\While{$|\mathcal{V}| \leq n_\text{nodes}$}{
$v_k \gets \text{RAND}(\mathcal{V})$

$\mu_q \gets \text{RANDMEANAROUND}(v_k)$

$(\text{status}, \Sigma_q, e_{k \to q}) \gets \Pi(\mu_k, \Sigma_k, \mu_q, N)$

\If{$\text{status} == \text{success}$} {
$\mathcal{V} \gets \mathcal{V} \cup \{(\mu_q, \Sigma_q)\}$

$\mathcal{E} \gets \mathcal{E} \cup \{e_{k \to q})\}$
}
}
\end{algorithm}
\vspace{-0.8cm}
\begin{algorithm}
\label{alg: edge_rewiring}
\caption{Belief roadmap construction with edge rewiring (REVISE)}
$\mathcal{V} \gets \{(\mu_\mathcal{I}, \Sigma_\mathcal{I})\}, \mathcal{E} \gets \emptyset$

$\mathcal{V}_\text{sample} \gets \{(\mu_\mathcal{I}, \Sigma_\mathcal{I})\}$

\While{$|\mathcal{V}| \leq n_\text{nodes}$}{
$v_{k, \text{spl}} \gets \text{RAND}(\mathcal{V}_\text{sample})$

$\mu_q \gets \text{RANDMEANAROUND}(v_{k, \text{spl}})$

$(\text{status}, \Sigma_{q, \text{spl}}, \_) \gets \Pi(\mu_{k, \text{spl}}, \Sigma_{k, \text{spl}}, \mu_q, N)$

\If{$\text{status} == \text{success}$} {
$\mathcal{V}_{\text{sample}} \gets \mathcal{V}_{\text{sample}} \cup \{(\mu_q, \Sigma_q \}$

$\Sigma_k \gets \text{COV-LOOKUP}(\mu_k, \mathcal{V})$

$(\_, \Sigma_q, e_{k \to q}) \gets \Pi(\mu_k, \Sigma_k, \mu_q, N)$

$\mathcal{V}_{\text{near}} \gets \text{NEAR}(\mu_q)$

$\Sigma_{q_{\min}} \gets \Sigma_q,\ e_{\to q} \gets e_{k \to q}$

\For{$v_i \in \mathcal{V}_{\text{near}}$}{
$(\text{status}, \Sigma_{q_i}, e_{i \to q}) \gets \Pi(\mu_i, \Sigma_i, \mu_q, N)$

\If{$\lambda_{\max}(\Sigma_{q_i}) \leq \lambda_{\max}(\Sigma_{q_{\min}})$}{
$\Sigma_{q_{\min}} \gets \Sigma_{q_i}, \ e_{\to q} \gets e_{i \to q}$

}
}
$\mathcal{V} \gets \mathcal{V} \cup \{(\mu_q, \Sigma_{q_{\min}})\}$

$\mathcal{E} \gets \mathcal{E} \cup \{e_{\to q})\}$

$\mathcal{V}_{\text{near}} \gets \text{NEAR-EXCEPT-ANCESTORS}(\mu_q)$

\For{$v_i \in \mathcal{V}_{\text{near}}$}{

$(\text{status}, \Sigma_{i_q}, e_{q \to i}) \gets \Pi(\mu_q, \Sigma_q, \mu_i, N)$

\If{$\lambda_{\max}(\Sigma_{i_q}) \leq \lambda_{\max}(\Sigma_{i})$}{

$\mathcal{V} \gets (\mathcal{V} \setminus \{(\mu_i, \Sigma_{i})\}) \cup \{(\mu_i, \Sigma_{i_q})\}$

$\mathcal{E} \gets (\mathcal{E} \setminus \{e_{\to i})\}) \cup \{e_{q \to i})\}$

$\text{RECOMPUTE-DESCENDANTS}(v_i)$

}
}
}
}
\end{algorithm}

REVISE depends on a belief roadmap construction procedure with an edge rewiring step similar to that of RRT* \cite{karaman2011incremental}. In this procedure (see Algorithm \ref{alg: edge_rewiring}), the \textit{minimum-cost} edge is added to the query mean $\mu_q$. Then, we check the nodes that neighbor the new node to see if steering via $\mu_q$ lowers the cost to reach those nodes. If so, we revise edges in the tree to maintain \textit{minimum-cost} paths to neighbors of the new node. We also recursively recompute costs of edges from revised nodes to their descendants, propagating lower costs down to leaf nodes, as in CC-RRT* \cite{luders2013robust}. 
Because new nodes are added based on $\mathcal{V}_{\text{sample}}$, which is equivalent to the set of nodes that would be added to the tree by Algorithm \ref{alg: no_rewiring}, if Algorithms \ref{alg: no_rewiring} and \ref{alg: edge_rewiring} are run with the same random seed, they will produce trees with the same node means, but the tree generated by Algorithm \ref{alg: edge_rewiring} will have equal or smaller node covariances.

Algorithms \ref{alg: no_rewiring} and \ref{alg: edge_rewiring} can use any valid edge controller $\Pi$. When $\Pi =$ Problem \ref{prob: min_cov_grf}, Algorithm \ref{alg: edge_rewiring} leads to coverage provably equal to or better than that of Algorithm \ref{alg: no_rewiring}, even though the objective of Problem \ref{prob: min_cov_grf} is not an admissible objective for RRT*. This property also holds under mild conditions when $\Pi =$ Problem \ref{prob: robust_cov_grf}. This property is formalized in Theorem \ref{thm: min_cov_coverage}, with a proof sketch in Appendix \ref{app: coverage_proofs}.
\begin{theorem} \label{thm: min_cov_coverage}
Consider two belief roadmaps $\mathcal{T}(\mathcal{I}, N, n_{\text{nodes}})$ and $\mathcal{T}^*(\mathcal{I}, N, n_{\text{nodes}})$, such that $\mathcal{T}$ is generated by Algorithm \ref{alg: no_rewiring} and $\mathcal{T}^*$ is generated by Algorithm \ref{alg: edge_rewiring}. Suppose both roadmaps are constructed with $\Pi =$ Problem \ref{prob: min_cov_grf}, or with $\Pi =$ Problem \ref{prob: robust_cov_grf} under the conditions that (1) for all $i$ and all $j$ and any initial covariance $\Sigma_\mathcal{I}$, $\lambda_{j}(S^{(i)})$ is a monotonically increasing function of $\lambda_{j}(\Sigma_{\mathcal{I}})$, where $\lambda_j(\cdot)$ is equal to the $j$th largest eigenvalue of a matrix, and (2) for all $\Sigma_\mathcal{I}$, $\sum_{i=1}^{4n} (\overline{\mathbf{W}}^{(i)}-\overline{\mathbf{W}}) \propto \mathbf{k}$, where $k$ is a constant vector. For any goal distribution $\mathcal{G}$, $\mathbb{P}(\mathcal{G} \text{ reachable from } \mathcal{T}) \leq \mathbb{P}(\mathcal{G} \text{ reachable from } \mathcal{T}^*)$.
\end{theorem}
\section{EXPERIMENTS}\label{sec: experiments}
We illustrate the benefits of REVISE with multi-query and single-query motion planning experiments for a quadrotor in a 2D plane navigating through spatially correlated wind. The quadrotor dynamics are modeled as a triple integrator with the state-space dynamics given by \cref{eq:dynamics} with
$$
A = \begin{bmatrix} I_2 & \Delta t I_2 & \frac{\Delta t^2}{2}I_2 \\ 0_2 & I_2 & \Delta t I_2 \\ 0_2 & 0_2 & I_2 \end{bmatrix}, B = \begin{bmatrix} 0_2 \\ 0_2 \\ \Delta t I_2 \end{bmatrix}, G = \begin{bmatrix} \Delta t I_2 \\ 0_2 \\ 0_2\end{bmatrix}
$$
and with $\Psi(\phi(\mathbf{x}))$ representing the wind field. 

For each set of experiments, we present an ablation study. We construct a belief roadmap using REVISE (Algorithm \ref{alg: edge_rewiring} with $\Pi =$ Problem \ref{prob: robust_cov_grf}). We also generate a roadmap using a baseline algorithm (Algorithm \ref{alg: no_rewiring} with $\Pi =$ Problem~\ref{prob: min_cov_grf}) \cite{aggarwal2024sdp, ridderhof2022chance} . Finally, we generate a ``robust ablation'' roadmap with Algorithm~\ref{alg: no_rewiring} and $\Pi =$ Problem~\ref{prob: robust_cov_grf}, and a ``rewired ablation'' roadmap with Algorithm~\ref{alg: edge_rewiring} and $\Pi = $ Problem~\ref{prob: min_cov_grf}.

We sample the wind field at a frequency of 1 m over a 121 $\text{m}^2$ space, with $x, y \in [0, 10]$, and use bilinear interpolation to extrapolate to other points in the state space. The mean wind field is a counterclockwise flow given by $[(5-y)/4, (x-5)/4]$ m/s at each sampled point. The correlation coefficient between two sampled points $(x_1, y_1)$ and $(x_2, y_2)$ is given by $\max\left(0, 0.3-\frac{\sqrt{(x_1 - x_2)^2 + (y_1 - y_2)^2}}{10 \sqrt{2}}\right)$ $\text{m}^2/\text{s}^2$. For the multi-query experiment, the wind variance is $0.2$ $\text{m}^2/\text{s}^2$ at all sample points, and for the single-query experiment the wind variance is $0.2$ $\text{m}^2/\text{s}^2$ at all sample points except in a high-variance box bounded by $x, y \in [3, 7]$, where it is 6 $\text{m}^2/\text{s}^2$. 

We use $N = 6$ for all experiments, with $\Delta t = 0.1$ for the multi-query and $\Delta t= 0.2$ for the single-query experiments. We constrain the quadrotor position to $ [0, 10]$ m, velocity to $[-10, 10]$ m/s, and acceleration to $[-100, 100]$ m/$\text{s}^2$.
\subsection{Multi-query Experiment}
We construct four different belief roadmaps, using the baseline algorithm, robust ablation, rewired ablation, and REVISE. Each roadmap has 500 nodes and $\mathcal{I} = ([5, 5, 0, 0, 0, 0], 0.1I)$. Then, we sample 100 random goal means that are reachable from all four roadmaps, and find plans to each goal for each roadmap. We use 200 Monte Carlo simulations for each goal to evaluate the accuracy of each plan, and evaluate the Wasserstein distance between the planned and actual final distribution.

\begin{table}[h]
\vspace{-0.2cm}
\caption{Minimum, median, and maximum Wasserstein distance between the planned and actual final distribution for each roadmap over 100 random goals.}
\centering
\begin{tabular}{|c |c |c |c|} 
 \hline
  & Min $W_2$ & Median $W_2$ & Max $W_2$ \\ [0.5ex] 
 \hline\hline
 Baseline \cite{ridderhof2022chance, aggarwal2024sdp} & 0.0013 & 0.178 & 82.7 \\ 
 \hline
 Robust ablation & 0.0045 & 0.0359 & 7.38 \\
 \hline
 Rewired ablation & \textbf{0.0008} & 0.0302 & 308.5 \\
 \hline
 REVISE (ours) & 0.0028 & \textbf{0.0176} & \textbf{7.08} \\
 \hline
\end{tabular}
\vspace{-0.2cm}
\label{tab: multiquery_mse}
\end{table}
 \begin{figure}[thpb]
      \centering
      \includegraphics[scale=0.5]{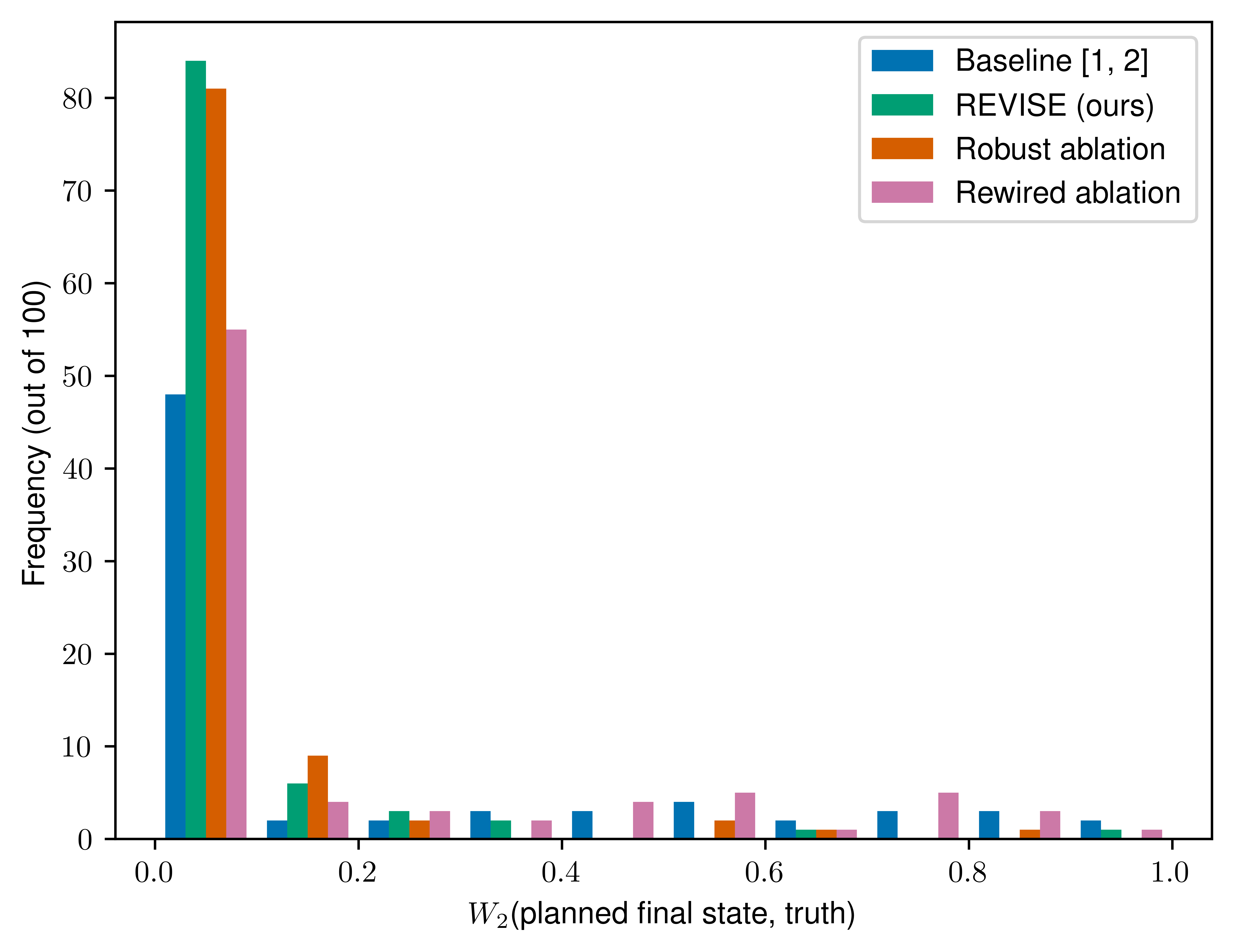}
      \vspace{-0.4cm}
      \caption{Distribution of Wasserstein distance between the planned and actual final distribution for 100 different goals reachable from the multi-query roadmap. Trials with $W_2$(plan, goal) $> 1$ are not shown.}
      \label{fig: multi_query_results}
\end{figure}

REVISE demonstrates lower median final $W_2$(plan, goal) than the rewired ablation, and the robust ablation demonstrates lower median final $W_2$(plan, goal) than the baseline, as seen in Figure \ref{fig: multi_query_results} and Table \ref{tab: multiquery_mse}. This suggests that our robust covariance steering algorithm improves roadmap accuracy, leading to lower divergence from the planned trajectory. REVISE has lower median $W_2$(plan, goal) than the robust ablation and the rewired ablation has lower median $W_2$(plan, goal) than the baseline. This can be interpreted as edge rewiring generating smooth plans that are easy to follow.

\subsection{Single-query Experiment}
\begin{figure}[htpb]
      \centering
      \includegraphics[width=\columnwidth]{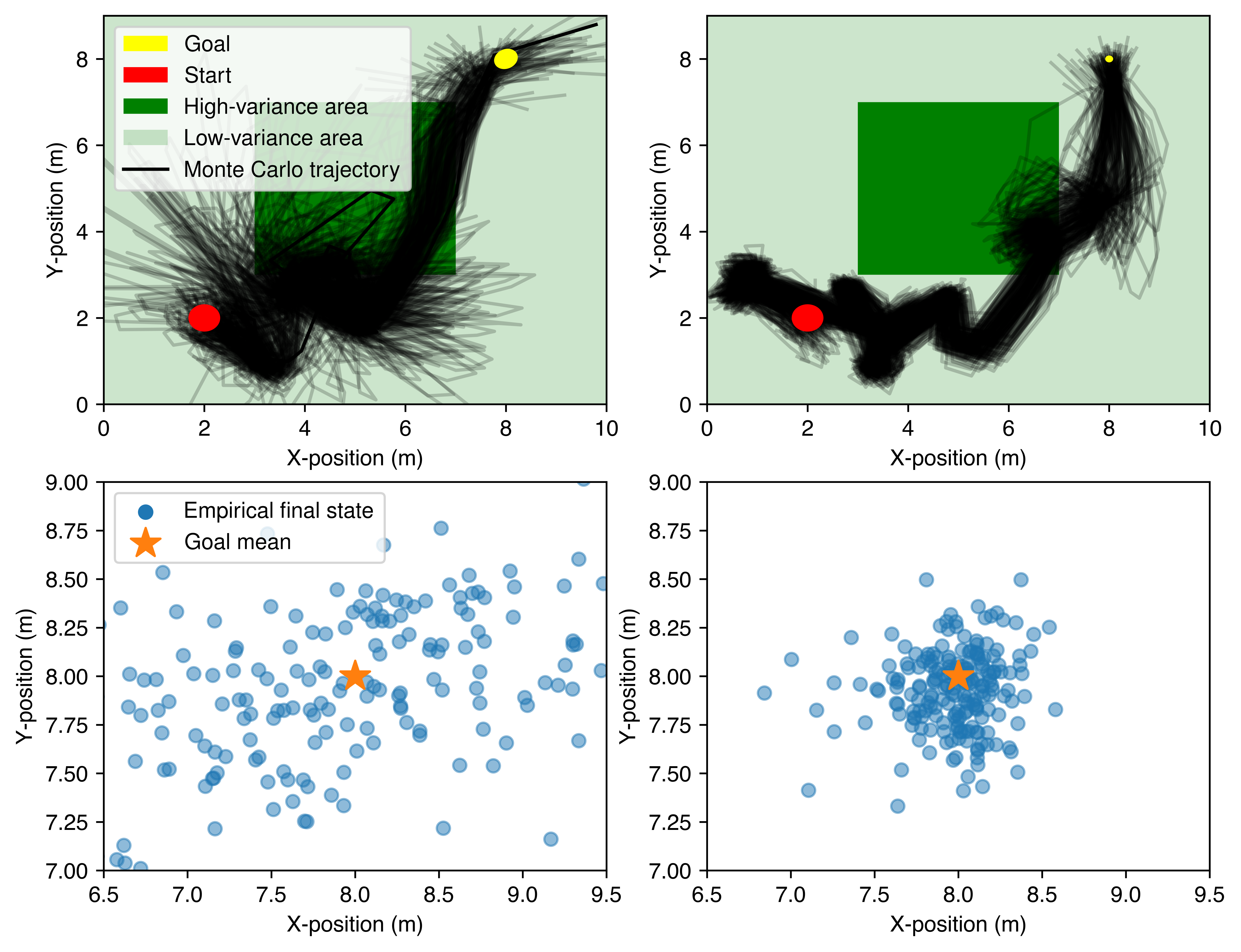}
      \vspace{-0.8cm}
      \caption{Trajectories and final state distribution for the single query experiment. Trial 1 (with random seed 0) plotted for each method. Top left: Baseline trajectories, top right: REVISE trajectories, bottom left: baseline final state distribution, bottom right: REVISE final state distribution.}
      \label{fig: single_query_results}
\end{figure}

We construct 80 different belief roadmaps, using the baseline algorithm, robust ablation, rewired ablation, and REVISE with 20 random seeds. Each roadmap has 200 nodes, $\mathcal{I} = ([2, 2, 0, 0, 0, 0], 0.1I)$, and $\mathcal{G} = ([8, 8, 0, 0, 0, 0], 0.2I)$. We adapted our algorithms to single-query planning by checking for feasible paths to the goal distribution $\mathcal{G}$ during roadmap construction. For the rewired ablation and REVISE, whenever a node $v_\text{new}$ is added to the belief roadmap, we check if it is feasible to steer from $v_\text{new}$ to $\mathcal{G}$. For the baseline and robust ablation, we stop checking for paths to the goal as soon as a feasible path is found. We employ Monte Carlo simulations to evaluate the quality of the plans found by each algorithm. We calculate the median final mean squared error, Wasserstein distance between the planned and actual final state distribution, and plan cost (largest eigenvalue of the planned covariance at the goal node) for each roadmap, over all 20 trials. Trajectories generated by the baseline and by REVISE are visualized in Figure \ref{fig: single_query_results}.

\begin{table}[h]
\caption{Median Wasserstein distance between planned and actual final distribution, final MSE, and largest eigenvalue of planned covariance at goal across 20 trials for the single-query experiment.}
\begin{tabular}{|c |c |c |c|} 
 \hline
  & MSE & $W_2$(plan, truth) & Planned $\lambda_{\max}(\Sigma_\mathcal{G})$ \\ [0.5ex] 
 \hline\hline
 Baseline \cite{ridderhof2022chance, aggarwal2024sdp} & 13.48 & 12.17 & 0.027 \\ 
 \hline
 Robust ablation & 0.121 & 0.118 & 0.026 \\
 \hline
 Rewired ablation & 0.099 & 0.054 & \textbf{0.004} \\
 \hline
 REVISE (ours) & \textbf{0.066} & \textbf{0.049} & 0.010 \\
 \hline
\end{tabular}
\label{tab: single_query_results}
\end{table}
Roadmaps generated with REVISE and the rewired ablation have lower planned $\lambda_{\max}(\Sigma_\mathcal{G})$ than roadmaps generated with the baseline and the robust ablation. Furthermore, the robust ablation failed to reach the goal in 1 trial (out of 20), while REVISE always successfully reaches the goal node. This demonstrates empirically that edge rewiring leads to lower-cost plans, which is consistent with Theorem \ref{thm: min_cov_coverage}.  Our robust covariance steering algorithm improves plan accuracy, as demonstrated by REVISE outperforming the rewired ablation and the robust ablation outperforming the baseline on MSE and $W_2$(plan, goal). Finally, the rewired ablation has the lowest median planned $\lambda_{\max}(\Sigma_\mathcal{G})$. This is expected, because edge rewiring lowers plan costs but the robust objective increases plan costs due to its conservative approximation of the final state covariance, which is one potential downside of REVISE.
\section{CONCLUSION}
We introduced REVISE, a multi-query algorithm for forward probabilistic planning in a Gaussian random field. Our two main contributions were a novel robust edge controller for covariance steering in a Gaussian random field, and a new procedure for revising belief roadmap edges during construction that provably improves coverage. REVISE is supported by theoretical analysis and experiments on a 6DoF model. In the future, we plan to incorporate robust constraints into REVISE, which will likely improve constraint satisfaction but greatly increase computation time.



\section*{APPENDIX}
\subsection{Proof Sketch of Theorem \ref{thm: min_cov_coverage}}\label{app: coverage_proofs}
We state two lemmas used to prove Theorem \ref{thm: min_cov_coverage}.
\begin{lemma}\label{lemma:funnel}
For initial state distributions $\mathcal{I}_1 = \mathcal{N}(\mu_{\mathcal{I}_1}, \Sigma_{\mathcal{I}_1})$, $\mathcal{I}_2= \mathcal{N}(\mu_{\mathcal{I}_2}, \Sigma_{\mathcal{I}_2}),$ suppose $\Sigma_{\mathcal{I}_2} \succeq \Sigma_{\mathcal{I}_1}$. Suppose $\Pi$ is used to steer to a goal mean $\mu_\mathcal{G}$, with $\Pi$ = Problem \ref{prob: min_cov_grf} or $\Pi$ = Problem \ref{prob: robust_cov_grf} under conditions (1) and (2) from Theorem \ref{thm: min_cov_coverage}. Then, steering from distribution $\mathcal{I}_1$ yields final state covariance $\Sigma_{\mathcal{G}_1}$, and steering from $\mathcal{I}_2$ yields $\Sigma_{\mathcal{G}_2}$, with $\Sigma_{\mathcal{G}_2} \succeq \Sigma_{\mathcal{G}_1}$.
\end{lemma} 
\begin{lemma} \label{lemma: same_means}
Consider two trees $\mathcal{T}_1 = (\mathcal{V}_1, \mathcal{E}_1)$ and $\mathcal{T}_2= (\mathcal{V}_2, \mathcal{E}_2)$, such that $\mathcal{T}_1$ is built with Algorithm \ref{alg: no_rewiring} and $\mathcal{T}_2$ is built with Algorithm \ref{alg: edge_rewiring}. Suppose both trees are built with the same random seed, input distribution $\mathcal{I}$, maneuver length $N$, and number of nodes $n_\text{nodes}$, with $\Pi =$ Problem \ref{prob: min_cov_grf} or $\Pi$ = Problem \ref{prob: robust_cov_grf} under conditions (1) and (2) from Theorem \ref{thm: min_cov_coverage}. Then, $\mathcal{T}_1$ and $\mathcal{T}_2$ will have the same set of node means, and for any node pair $v^{(\mathcal{T}_1)}$, $v^{(\mathcal{T}_2)}$ such that $v^{(\mathcal{T}_1)} \in \mathcal{V}_1$, $v^{(\mathcal{T}_2)} \in \mathcal{V}_2$ and 
$\overline{v^{(\mathcal{T}_1)}} = \overline{v^{(\mathcal{T}_2)}}$, $\Sigma_{v^{(\mathcal{T}_1)}} \succeq \Sigma_{v^{(\mathcal{T}_2)}}$.
\end{lemma}

\textbf{\textit{Proof Sketch of Theorem \ref{thm: min_cov_coverage}}}:
Suppose $\mathbb{V}(\mathcal{I}, N, n_\text{nodes})$ is the set of tuples of ordered node means, where each tuple corresponds to a particular tree generated by Algorithm \ref{alg: no_rewiring} with $\Pi =$ Problem \ref{prob: min_cov_grf}. $\mathbb{V}^*(\mathcal{I}, N, n_\text{nodes})$ is defined the same way, but for Algorithm \ref{alg: edge_rewiring}. By Lemma \ref{lemma: same_means}, these sets are equal. Both sets are Borel-measurable spaces, so we construct a probability space $(\mathbb{V}, \mathcal{B}(\mathbb{V}), \mu)$, where $\mu$ measures the probability that $\mathcal{V} \in \mathbb{V}$ is generated by Algorithm \ref{alg: no_rewiring}, and $\mathcal{B}(\mathbb{V})$ is the Borel set of $\mathbb{V}$. $\mathbb{V}_{\text{reach}}(\mathcal{I}, N, n_\text{nodes})$ is the set of tuples of ordered node means in $\mathbb{V}(\mathcal{I}, N, n_\text{nodes})$ corresponding to trees that can reach a goal distribution $\mathcal{G}$. We also construct a corresponding probability space $(\mathbb{V}^*, \mathcal{B}(\mathbb{V}^*), \mu) = (\mathbb{V}, \mathcal{B}(\mathbb{V}), \mu)$ and set $\mathbb{V}^*_{\text{reach}}(\mathcal{I}, N, n_\text{nodes})$ for Algorithm \ref{alg: edge_rewiring}. By Lemma \ref{lemma:funnel} and the equality of $\mathbb{V}$ and $\mathbb{V}^*$, $\mathbb{V}_\text{reach} \subseteq \mathbb{V}^*_\text{reach}$. Then, if $\mathcal{T}$ is generated by Algorithm \ref{alg: no_rewiring} with corresponding tuple of ordered node means $\mathcal{V}$ and $\mathcal{T}^*$ is generated by Algorithm \ref{alg: edge_rewiring} with corresponding tuple of ordered node means $\mathcal{V}^*$, then for any $\mathcal{G}$, $\mathbb{P}(\mathcal{V} \in \mathbb{V}_\text{reach}) \leq \mathbb{P}(\mathcal{V}^* \in \mathbb{V}^*_\text{reach})$, so $\mathbb{P}(\mathcal{G} \text{ reachable from } \mathcal{T}) \leq \mathbb{P}(\mathcal{G} \text{ reachable from } \mathcal{T}^*)$.

\addtolength{\textheight}{-12cm}   


\bibliographystyle{IEEEtran}
\bibliography{IEEEabrv,ref}

\end{document}